\documentclass[10pt,twocolumn,letterpaper]{article}

\usepackage{cvpr}
\usepackage{times}
\usepackage{epsfig}
\usepackage{graphicx}
\usepackage{amsmath}
\usepackage{amssymb}
\usepackage[utf8]{inputenc}

\usepackage[pagebackref=true,breaklinks=true,letterpaper=true,colorlinks,bookmarks=false]{hyperref}

 \cvprfinalcopy 

\def\cvprPaperID{2354} 

\ifcvprfinal\fi
\begin{document}

\title{Deep Roto-Translation Scattering for Object Classification}

\author{Edouard Oyallon and St\'ephane Mallat\\
D\'epartement Informatique, Ecole Normale Sup\'erieure \\
45 rue d'Ulm, 75005 Paris\\
{\tt\small edouard.oyallon@ens.fr}
}


\maketitle

%
\begin{abstract}
Dictionary learning algorithms or supervised deep convolution networks have
considerably improved the efficiency of predefined feature representations such
as SIFT. We introduce a deep scattering convolution network, 
with complex wavelet
filters over spatial and angular variables. This representation 
brings an important improvement to
results previously obtained with predefined features over object image
databases such as Caltech and CIFAR. The resulting accuracy
is comparable to results obtained 
with unsupervised deep learning and dictionary based representations.
This shows that refining image representations by using geometric 
priors is a promising direction to improve image classification and 
its understanding.
\end{abstract}

\section{Introduction}
Learning image representations has considerably enhanced
image classification results compared to geometric 
features such as edge descriptors, or, SIFT and HOG \cite{Lowe:04,Dalal2005} patch representations.
Learning may thus seem to be a more promising direction for
improving image analysis rather than refining geometric image analysis. 
This paper aims at showing that 
understanding how to take
advantage of geometrical image properties can define image representations,
providing
competitive results with state of the art unsupervised
learning algorithms. It shows that refining
geometric image understanding remains highly promising
for image classification. 

Supervised deep neural network learning achieves state-of-the-art results
on many databases \cite{conf/nips/KrizhevskySH12,lee2014deeply}. However, several works \cite{journals/corr/ZeilerF13,journals/corr/GirshickDDM13}
have shown that the Alex-net \cite{conf/nips/KrizhevskySH12} trained on ImageNet still performs
very well on different databases such as Caltech or PASCAL VOC. The output
of this neural network can thus be 
considered as a ``super SIFT'' image descriptor, 
which is used as an input to a linear SVM classifier \cite{journals/corr/ZeilerF13,journals/corr/GirshickDDM13}.
It indicates that this deep network is capturing important generic
image properties, which are not dependent upon the classes used for
training. In the same spirit, 
unsupervised deep learning \cite{conf/icml/LeRMDCCDN12} as well 
as unsupervised bag of words \cite{citeulike:1321552} or dictionary learning with spatial pyramid\cite{conf/cvpr/LazebnikSP06}
have improved classification 
results previously obtained with engineered feature vectors 
such as SIFT or HOG, 
on complex object recognition databases.
However, these unsupervised learning algorithms are tailored
to each databases. One may wonder whether their 
improved performances
result from an adaptation to the specific properties of each databases, or
whether these unsupervised representations 
capture refined geometric image properties
compared to SIFT or HOG features. 

A scattering convolution network is constructed with
predefined complex wavelet filters, which are adapted to geometric image variabilities
\cite{journals/corr/abs-1101-2286}.
It provides a mathematical and algorithmic framework to incorporate refined
geometric image priors within the representation.
Since images are projections of 3D scenes under various view points, 
the main source of geometric image variabilities comes from 
rigid movements, and deformations
resulting from perspective projections. An important issue is to 
build adaptive invariants 
to these sources of variability, which preserve essential
information to discriminate different classes. 
A translation invariant scattering network was studied in
\cite{journals/pami/BrunaM13} for digit image classification and texture recognition, but which
was not powerful enough to classify complex objects as in Caltech or CIFAR.
A translation and
rotation invariant deep scattering network was introduced in \cite{conf/cvpr/SifreM13}
to classify textures with strong rotations and scalings. However, imposing
rotation invariance is a prior which is too strong for image object and 
scene classifications, which are typically not fully rotation invariant.

Section \ref{scatnet} introduces a scattering representation which is
translation invariant, and which
efficiently represents rotation variability without
imposing full rotation invariance. It yields a representation which
complements SIFT type coefficients, with coefficients incorporating 
interactions between scales and angles. 
This roto-translation scattering representation is nearly complete in the
sense that good quality images can be recovered from roto-translation
scattering coefficients \cite{joanreco}. It is also stable to additive
perturbations and small deformations, which guarantees to avoid the
type of instability observed in some deep networks \cite{journals/corr/SzegedyZSBEGF13}. 
In this architecture, the loss of information
only appears at the final supervised classification stage, which computes
invariants adapted to the classification task. It 
includes an orthogonal least square
supervised feature selection followed
by a linear or a Gaussian kernel SVM. 

This scattering representation is tested in Section \ref{numeraics}
over Caltech and CIFAR data bases for object classification. 
It yields results which are well above all other representation which do not
incorporate any learning, based on SIFT type features or
with random weight deep networks. 
It also gives competitive results with state of the art
unsupervised learning procedure adapted to each databases, which
indicates that these unsupervised learning algorithm do not
capture geometric transformations which are more powerful than rigid movements
and small deformations. Computations can be reproduced with a software that is available at {\it http://www.di.ens.fr/data/software}.


\section{Roto-Translation Scattering Networks}
\label{scatnet}

Images have important geometric variability due to perspective projections
of 3D scenes under various viewpoints. It includes a combination of rigid
movements and deformations. This section introduces a separable scattering
transform, which constructs a nearly complete representation, based on
elementary features, which linearizes important geometric variability.
This representation is used for object classification. 
Scattering networks are particular classes of convolution 
networks \cite{journals/corr/abs-1101-2286}, whose filters are computed with wavelets. 
They are introduced in the framework of convolution networks to better
understand the specificities of their architecture.

\subsection{Convolution Network Cascade}

A convolutional network is a multilayer architecture, which cascades
spatial convolutions and pooling operators, which
includes sub-samplings \cite{conf/iscas/LeCunKF10}. 
These networks compute progressively
more invariant image descriptors over multiple layers indexed by 
$0 \leq j \leq J$.

\begin{figure*}
\center
\includegraphics[natwidth=2280,natheight=1450,width=\textwidth]{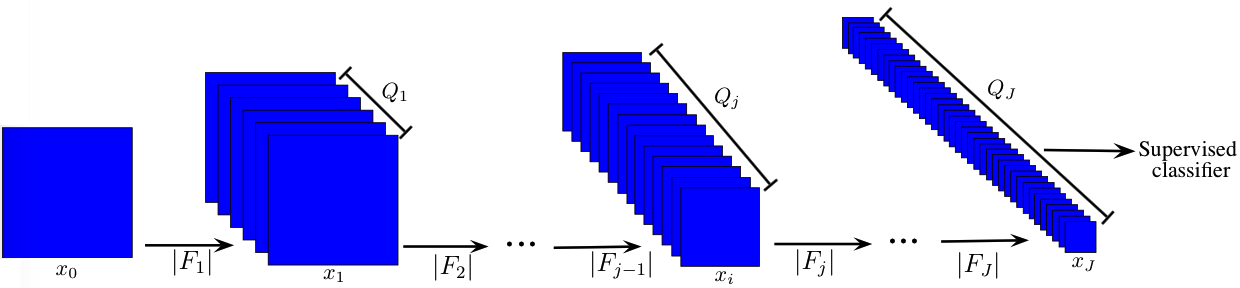}
\caption{A convolution network computes a layer $x_j$ by applying
a linear operator $F_j$ to $x_{j-1}$, 
followed by a non-linearity which we choose to be
a modulus.}
\label{figure1}
\end{figure*}

For $j=0$, the network illustrated in Figure \ref{figure1}
takes in input an image $x$ of $P$ pixels, with
potentially $Q_0 = 3$ color frames. For $j > 0$, each layer 
$x_j (p,q)$ is a set of $Q_j$ image frames, which correspond to
different ``feature types'' indexed by $1 \leq q \leq Q_j$.
Each feature image has $P_j$ pixels indexed by $p$.
It is computed from $x_{j-1}$ by applying a linear operator $F_j$
to $x_{j-1}$, followed by a non-linearity, which may be a rectifier, a
thresholding, a modulus or some other non-linearities ``pooling'' functions
\cite{conf/iscas/LeCunKF10}. Convolution networks impose that for each fixed $q$,
$F_{j}$ computes a convolution of $x_{j-1} (p,q)$ along $p$, 
with a filter which depends upon $q$. The
operator $F_j$ also linearly combines the $Q_{j-1}$ image 
frames of $x_{j-1}$ indexed by $q$. 
The output vector $F_j x_{j-1}$ is then transformed by a non-linear
``pooling operator'' which may incorporate a rectifier, a modulus,
a thresholding \cite{conf/iscas/LeCunKF10}.
For scattering transforms, $F_j x_{j-1}$ is a complex valued signal and the 
non-linearity is a complex modulus, so we write:
\begin{equation}
\label{eqnadf}
x_{j} = | F_j |\, x_{j-1}~.
\end{equation}
Figure 1 illustrates this computational architecture introduced by LeCun
\cite{conf/iscas/LeCunKF10}. 

The operators $|F_j|$ progressively propagate $x$ across the network
until the last layer $x_J$. 
The cascade of convolutions produce operators of progressively
wider supports as $j$ increase. The depth $j$ thus corresponds
to a scale index of the non-linear network features.
A classifier is applied to the output $x_J$.
It may be a linear SVM, a RBF network, or some other
fully connected double layer classification networks \cite{conf/nips/KrizhevskySH12,conf/iscas/LeCunKF10}. 
In our numerical experiments, we use a dimensionality reduction step followed by a Gaussian SVM.

The network architecture is specified by the dimensions $P_j \times Q_j$ of
each layer and by the non-linear pooling operator. This is a delicate
step, which is usually done through an ad-hoc engineering 
trial and error process. 
Given this architecture, one must then optimize each operators $F_j$
to achieve a low classification error. 
Experiments
have been performed with random weights \cite{conf/icml/SaxeKCBSN11}. Better results are however
obtained with unsupervised training of the weights, using auto-encoders \cite{conf/icml/VincentLBM08}. When enough labeled examples are available, even better results are
obtained with supervised training algorithms which back-propagates 
classification errors \cite{LeCun:89}. 
The Alex-Net is an example of supervised deep network trained with labeled
images of the ImageNet data basis \cite{conf/nips/KrizhevskySH12}.

\subsection{Scattering Network}

A scattering network is a convolutional network whose architecture
and filters are not learned, but are predefined wavelets.
These wavelets are adapted to the type of geometric
invariants and linearization that need to be computed. Image
classification typically requires to build
image features which are locally invariant to translations and stable to
deformations. It should thus linearize small deformations
so that these deformations can be taken into account or removed
with a linear operator, which is adjusted by the final supervised classifier. 

Scattering networks compute inner-layer coefficients $x_j(p,q)$,
which are nearly invariant to translations 
of the input image $x$ by less than $2^j$.
Each image frame is sub-sampled at intervals $2^{j-1}$. The factor $2$ 
oversampling avoids aliasing phenomena. 
If $x$ has $P$ pixels then 
each network layer $x_j$ has $Q_j$ frames of $P_j = P\, 2^{-2j+2}$ pixels. 
The size $Q_j$ does not result from an ad-hoc decision but
depends upon the choice of geometric invariant as we shall see.

It has been proved mathematically that translation
invariance and linearization of deformations is obtained with wavelets 
\cite{journals/corr/abs-1101-2286}. They separate the image information along multiple
scales and orientations. Cascading wavelet transforms and modulus
non-linearities lead to translation invariant scattering
transforms \cite{journals/corr/abs-1101-2286}, which have been
applied to digit and texture classifications
\cite{journals/pami/BrunaM13}.
Rotation invariant scattering networks have been introduced 
by replacing wavelet spatial convolutions by convolutions along the
special Euclidean group of rigid movements. It takes into account both
translations and rotations \cite{conf/cvpr/SifreM13}. We introduce a simpler separable
convolution, which still has the ability to build invariants over 
rigid movements, but which leaves the choice of invariant to the 
final SVM classifier. 

Wavelet transforms can be computed with a cascade of linear filtering 
and sub-sampling operators,
which are called multi-rate filter banks \cite{Mallat1999}.
Although deep networks apply non-linearities at each layer,
they also include such linear cascades. Indeed, rectifier or modulus 
non-linearities have no impact over positive coefficients,
produced by averaging filters in the network. All non-linearities can
thus be removed from averaging filters output. Deep network 
computations can therefore
be factorized, 
as cascades of $j-1$ averaging and sub-sampling operators,
followed by a band-pass filter and a non-linearity, 
for multiple values of $j$. 
If the network includes a sub-sampling 
by a factor $2$ at each layer, then 
this is equivalent to a convolution with 
multiple wavelets of scale $2^{j}$, and a non-linearity.
These cascades are
followed by new cascades of $k-1$ averaging operators, for different
$k$, followed by a band-pass filter and a
non-linearity, and so on. This is  equivalent to 
convolutions with a second set of
wavelets of scale $2^{j+k}$, and a non-linearity.
The scales depend upon the number of averaging and sub-samplings 
along each network path,
and thus satisfy $1 \leq j + k \leq J$. 

In the following, we describe a second order
scattering transform operator $S_J$, which performs at most
two wavelet convolutions. 
The network output $x_J$ is computed with 
a first $2D$ spatial wavelet transform $W_1$ which performs spatial 
wavelet image convolutions whose phase are removed by a non-linear modulus.
We then apply a second wavelet transform $W_2$,
which is adapted to the desired invariants, not
only along translations but also along rotations. This is done
by computing separable
$2D$ convolutions with wavelets along space, and $1D$ convolutions
with wavelet along angle variables.
The output is averaged by an operator $A_J$ which performs
a spatial averaging at the scale $2^J$:
\[
x_J = S_J x = A_J \, |W_2|~ | W_1| x~.
\]
Higher order scattering transforms are obtained by cascading more wavelet
transforms, which can be adapted to other group of transformations. 
However, this paper concentrates on second order scattering along space
and rotation variable. This second order  should not
be confused with the network depth $J$,
which corresponds to the maximum spatial invariance scale $2^J$, 
and typically depends upon the image size. Next two sections describe the
implementations of 
the two wavelet transforms $W_1$ and $W_2$ and the averaging operator $A_J$.

\subsection{Spatial Wavelet Transform $W_1$}
The first wavelet transform $W_1$ separates the image component along different scales and orientations, 
by filtering the image $x$ with a family of wavelet $\psi_{j,\theta}$. 
These wavelets are obtained by
dilating by $2^j$ a mother wavelet $\psi(p)$, and 
rotating  its support with $r_\theta$ along $L$ angles $\theta$:
\[
\psi_{j,\theta} (p) = 2^{-2j} \, \psi(2^{-j} r_\theta p)~~\mbox{for}~~\theta = \frac {\ell \pi} L~.
\]
As in \cite{journals/pami/BrunaM13,conf/cvpr/SifreM13}, we choose a complex Morlet wavelet $\psi$ which is 
a Gaussian modulated by a complex exponential, to which is subtracted a
Gaussian to set its average to zero.
Figure 2 shows the real and imaginary parts of Morlet wavelets along $L=8$
angles. The modulus 
computes the envelop of complex wavelet coefficients, sub-sampled
at intervals $2^{j-1}$. Coefficients at the scale $2^j$ are stored at the depth
$j$:
\[
x_{j}^1 (p,\theta) = |x \star \psi_{j,\theta} (2^{j-1} p)|~.
\]
These coefficients are
nearly invariant to a translation of $x$ smaller than $2^j$. For $2^j < 2^J$, 
this invariance will be improved by further propagating these coefficients
up to layer $J$, 
with a second wavelet transform described in the next section.

\begin{figure}
\center
\includegraphics[natwidth=2280,natheight=1450,width=0.48\textwidth]{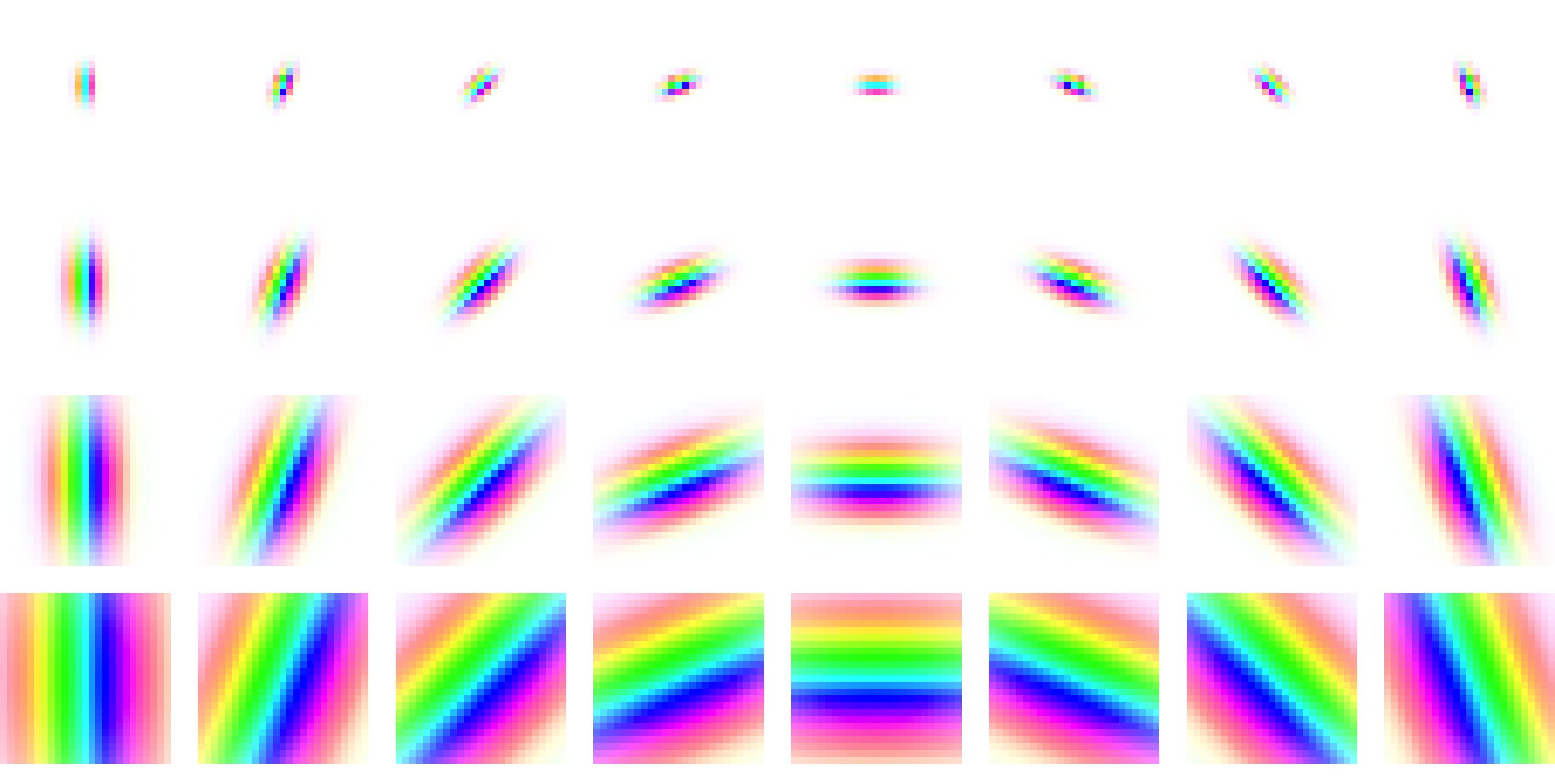}

\caption{Real and imaginary parts of Morlet wavelets at different scales
$2^j$ for $1 \leq j \leq 4$ and $L = 8$ orientations. Phase and amplitude are respectively given by the color and the contrast.(best viewed in color)}
\label{figure2}
\end{figure}

Coefficients at the scale $2^j$ correspond to deep network coefficients
of depth $j$ because in a deep network they
are calculated by cascading $j-1$
low-pass filters, and a final band-pass filter. The cascade of low-pass
filters defines a pyramid of low-passed images
$x \star \phi_j$, where $\phi_j$ is a scaled low-pass filter:
\[
\phi_j (p) = 2^{-2j} \phi(2^{-j} p). 
\]
Each 
$x \star \phi_j$ is computed by convolving
$x \star \phi_{j-1}$ with a low-pass filter $h$
followed by a sub-sampling. It is stored as an image indexed by $q=0$
in the layer $j$:
\begin{equation}
\label{aver}
x_j^1(p,0) = x \star \phi_j (p)  = h \star (x \star \phi_{j-1})(2 p) ~.
\end{equation}
Applying a modulus has no effect because these coefficients are positive.
The wavelet coefficients $x \star \psi_{j,\theta}$ are 
computed by applying a  complex band-pass filter $g_\theta$ 
followed by a sub-sampling. In this case, the modulus has a strong impact by
eliminating the complex phase. It is stored as 
an image indexed by $q=\theta$ in the layer $j$:
\begin{equation}
\label{diff2}
x_{j}^1 (p,\theta) = |x \star \psi_{j,\theta}(p)| = |g_\theta \star (x \star \phi_{j-1})(p)|.
\end{equation}

\begin{figure*}
\center
\includegraphics[natwidth=2280,natheight=1450,width=0.9\textwidth]{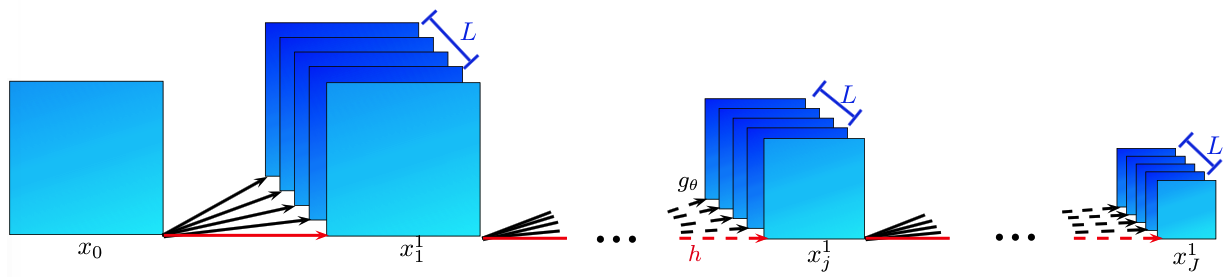}

\caption{A wavelet modulus $|W_1|$ computes averages
and modulus wavelet image frames at each layer $x_j^1$, by cascading
filtering, sub-sampling and modulus operators. }
\label{figure3}
\end{figure*}

The wavelet transform $W_1$ is thus implemented in a deep
network calculated with a cascade of
low-pass and band-pass filtering, followed by sub-samplings, illustrated in 
Figure \ref{figure3}.
Wavelet coefficients are computed at scales $2^j \leq 2^J$ and the lowest 
frequency image information is carried by the remaining averaged image
$x \star \phi_J$. 
The convolution cascades (\ref{aver}) and (\ref{diff2}) with $h$ and
$g_\theta$ can also be computed directly as convolutions with $\phi_j$
and $\psi_{j,\theta}$, using FFT's. For the sake of simplicity, we follow
this second approach and thus specify directly the $\phi$ and $\psi$
as opposed to the intermediate filters $h$ and $g_\theta$. We use a
Morlet wavelet $\psi$ and a Gaussian filter $\phi$,
further specified in \cite{journals/pami/BrunaM13}. The resulting
wavelet transform $W_1$ is a contractive linear operator,
which is nearly an isometry.

\subsection{Roto-Translation Wavelet Transform $W_2$}

Wavelet coefficients
$|x \star \psi_{j,\theta}|$ are translation invariant only up to the scale
$2^{j}$. Increasing this invariance up to $2^J$ means further propagating these
coefficients up to the last network layer $J$. This is be done by
applying a second wavelet transform $W_2$ which is now defined.
This second wavelet transform also recombines the output of wavelet filters
along different angles. It thus also measures the angular variability of
wavelet responses, as corner detectors. 

At a depth $j_1$, there are $\tilde Q_{j_1} = L$ wavelet image frames 
indexed by the angle $\theta = \ell \pi/L$ for $1 \leq \ell \leq L$:
\[
x^1_{j_1} (p,\theta) = |x \star \psi_{j_1,\theta} (2^{j_1-1} p)|.
\]
These coefficients are propagated to larger scales $2^{j}$
by computing convolutions and modulus with a new set of
spatial wavelets $\psi_{j,\theta} (p)$ at larger scales $2^j > 2^{j_1}$. 

As in deep convolution network architectures, we also
recombine the information in these image frames indexed by 
the angle $\theta$ in (\ref{diff2}).
To understand how to do so, let us compute the wavelet
coefficients of a rotated image $x_\alpha(p) = x(r_\alpha p)$ 
by an angle $\alpha$:
\[
x_\alpha \star \psi_{j_1,\theta} (p) = x \star \psi_{j_1,\theta-\alpha} (r_\alpha p)~.
\]
It rotates the spatial coordinates $p$ but also
 ``translates'' by $\alpha$ the angle parameter $\theta$.

Our goal is not to build a rotation invariant representation but
a representation which linearizes variabilites along
rotation angles. These totation variabilities
can thus be discriminated or removed by a linear classifiers at the output. 
We thus do not use a rotation invariant scattering
representation as in \cite{conf/cvpr/SifreM13}. 
To build a representation which 
is stable to rotations, and to deformations along rotations, we compute
a wavelet transform along the angle parameter $\theta$. It means
performing convolutions along $\theta$, with angular one-dimensional wavelets
$\overline \psi_{k} (\theta) = 2^{-k} \overline \psi( 2^{-k} \theta)$.
The resulting wavelet transform $W_2$
 computes separable convolutions along both the $2D$ spatial
variable $p$ and the angle variable $\theta$, with a $3D$ separable complex
wavelet defined by:
\[
\psi_{j,\beta,k} (p,\theta) = \psi_{j,\beta}(p) \,\overline \psi_{k} (\theta)~.
\]
It is a separable product of a spatial wavelet 
$\psi_{j,\beta}(p)$ of scale $2^{j}$ and 
an angular wavelet $\overline \psi_{k} (\theta)$ of scale $2^{k}$ 
for $1 \leq k \leq K < \log_2 L$. 
If $\overline \psi_k (\theta)$ are one-dimensional Morlet wavelets, then the
resulting separable wavelet transform $W_2$ is a stable and invertible 
operator, which nearly preserves the signal norm. 

The wavelet transform modulus for $j > j_1$ is computed
with a three-dimensional
separable convolution along the spatial and angular variables $(p,\theta)$, 
and it performs a sub-sampling along both variables. It has a spatial scale $2^j$
and is thus stored at the layer $j$, via an index $q$ which encodes $\theta$,
$\beta$, $j_1$ and the angular scale $2^k$:
\begin{equation}
\label{3donsdfsd}
x^2_{j}(p,q) = |x^1_{j_1} \star \psi_{j,\beta,k} (2^{-j-1} p,2^{-k-1} \theta) |~.
\end{equation}
It propagates $x^1_{j_1}$ towards network layers of depth $j > j_1$,
up to $j= J$. 
This 3D separable wavelet transform is either
computed with a cascade of filtering across the deep network layers,
or directly with $3D$ convolutions calculated with $FFT's$.

For $j < J$, we still need to propagate the second order
coefficients $x^2_j$ up to the largest spatial scale $2^J$. 
This could be done by applying a third wavelet transform $W_3$
which could also enforce more complex geometric invariants by recombining
information across angles and scales. In this implementation,
we directly apply a linear averaging $x^2_{j} \star \phi_J$
at the scale $2^J$. It averages each image frame of $x^2_{j}$ with a 
spatial convolution with $\phi_J (p) = 2^{-2J} \phi(2^{-J} p)$.

The last layer $x_J$ of this scattering network
is an aggregation of the image $x$, of first order wavelet 
modulus images $x^1_j$, and of second order coefficients
$x^2_j$ at all scales $2^j \leq 2^J$, 
all of them averaged at the scale $2^J$:
\[
x_J = S_J x = \Big\{ x \star \phi_J \,,\,x_{j}^1 \star \phi_J \,,\,
x_{j}^2 \star \phi_J \Big\}_{1 \leq j \leq J}~.
\]
First order coefficients $x_{j}^1 \star \phi_J$ are very similar
to SIFT \cite{Lowe:04} feature vectors. They provide information on average
energy distributions across scales and orientations over a neighborhood
of size $2^J$.
A scattering representations can thus be interpreted as an ``augmented''
SIFT representation with second order coefficients
$x_{j}^2 \star \phi_J$ providing information on interactions between
scales and angles in multi-scale neighborhoods.

This deep scattering is computed by cascading
the modulus $|W_1|$ of 
a first $2D$ spatial wavelet transform, followed by
the modulus $|W_2|$ of a second $3D$ separable wavelet transform 
along space and angles, followed by the averaging 
$A_J z = z \star \phi_J$:
\[
S_J x = A_J \,| W_2|~ | W_1|\, x~.
\]
Since $W_1$ and $W_2$ and $A_J$ are contractive operators 
it guarantees that $S_J$ is also contractive and hence stable to additive
perturbations. Moreover, since the wavelet transforms $W_1$ and $W_2$
and $A_J$ are Lipschitz stable relatively to deformations \cite{journals/corr/abs-1101-2286}, 
$S_J$ is also Lipschitz and hence linearizes small deformations. 
This guaranties to avoid the instabilities observed on deep networks
such as Alex-net \cite{journals/corr/SzegedyZSBEGF13} where a small image perturbation can
considerably modify the network output and hence the classification.

For images of $P$ pixels, each network layer $x_j$ has
$Q_j$ image frames of $P_j = P \,2^{-2j+2}$ pixels. 
For a gray level image such that $Q_0=1$, the resulting number of frames
are respectively $Q_1 = 9$, $Q_2=145$, $Q_3=409$, $Q_4=801$, $Q_5=1321$, $Q_6=1969$, and $Q_j=1+Lj+L^2j(j-1)\approx 64j^2$ when $j\gg 1$.
Color images are represented by the three Y,U,V color bands, and each
color band is decomposed independently. It thus multiplies the number
of image frames $Q_j$ by $3$ for all $0 \leq j \leq J$.
At the output layer $J$, the factor $2$ spatial oversampling is removed
so $P_J = P\, 2^{-2J}$. This last layer is thus
an aggregation of 
$3 P 2^{-2J}$ order $0$ coefficients in $x \star \phi_J$, 
plus $3 P 2^{-2J} L J$ order $1$ coefficients in the $x_j^1 \star \phi_J$,
and $3 P 2^{-2J} L^2 J(J-1)$ order $2$ coefficients in the $x_j^2 \star \phi_J$
arrays. For CalTech images of $P = 256^2$ pixels, decomposed with
$J = 6$ scales and $L = 8$ angles, there are $48$ order $0$ coefficients, 
$2\,10^3$ order $1$ coefficients, and $92\,10^3$ order $2$ coefficients.

Good quality images can be reconstructed from scattering coefficients
as long as the number of scattering coefficients is larger than the
number of image pixels \cite{joanreco}. 
Despite the invariance to translation, the roto-translation scattering
representation is thus nearly complete as long as 
$2^{-2J} L^2 J^2 \geq 1$. If $L=8$ then it is valid
for $J \geq 5$.

\section{Supervised Feature Selection}
\label{supervfea}

The scattering representation has a number of coefficients which is of the
same order as the original image. It provides a nearly complete signal
representation, which allows one to build a very rich set of
geometric invariants with linear projection operators. The choice of these
linear projection operators  is done at the supervised classification stage
with an SVM. Scattering coefficients are strongly correlated.
Results are improved by reducing the variance
of the representation, with a supervised feature selection, 
which considerably reduces the number of scattering coefficients
before computing an SVM classifier. This
is implemented with a supervised orthogonal least square regression \cite{chen1991orthogonal,blumensath2007difference}, which greedily selects
coefficients with a regression algorithm.

A logarithm non-linearity is applied to scattering coefficients in order
to separate low frequency
multiplicative components due to the variations of illuminations.
These low-frequency modulations add a constant to the logarithm of 
scattering coefficients which can then be removed with an appropriate linear
projector by the final classifier. Also, it linearizes exponential decay of the scattering coefficients across scales.

In the following, we denote by $\Phi^1 x = \{ \phi^1_p x \}_{p}$  the 
logarithm of scattering coefficients at a scale $2^J$. 
We are 
given a set of training images $\{x_i\}_i$ with their class label. 
The orthogonal least square selects a set of features 
adapted to each class $C$ with a linear 
regression of the one-versus-all indicator function
\[
f_C(x) = 
\left\{
\begin{array}{ll}
1 & \mbox{if $x$ belongs to class $C$}\\
0 & \mbox{otherwise}
\end{array}
\right.~.
\]
It iteratively selects a feature in the dictionary and updates the
dictionary.
Let
$\Phi^k x = \{ \phi^k_p x \}_{p}$ be  the dictionary at the $k^{th}$ iteration.
We select 
a feature $\phi^{k}_{p_{k}} x$, and we update the dictionary
by decorrelating all dictionary vectors, relatively
to this selected vector, over the training set $\{x_i \}_i$:
\[
\tilde \phi^{k+1}_p = \phi_p^k - \Big(\sum_i \phi^k_{p_k} (x_i)\, \phi^k_{p} (x_i)\Big) \phi_{p_{k}}^{k}.
\]
Each vector is then normalized
\[
\phi^{k+1}_p = \tilde \phi^{k+1}_p \Big(\sum_i |\tilde \phi^{k+1}_{p} (x_i)|^2\Big)^{-1}~.
\]
The $k^{th}$ feature $\phi^k_{p_{k}} x$ 
is selected so that the linear regression of 
$f_C (x)$ on $\{\phi^r_{p_{r}} x \}_{1 \leq r \leq  k}$ 
has a minimum mean-square error,
computed on the training set. This is equivalent to finding $\phi^k_{p_k}$ 
in $\Phi^k$ which maximizes the 
correlation $\sum_i f_C (x_i) \, \phi^k_{p} (x_i)$.

The orthogonal least square regression thus selects and computes
$K$ scattering  features $\{ \phi_{p_k} x \}_{k < K}$ for each class $C$,
which are linearly transformed into $K$ decorrelated and normalized  
features $\{ \phi^k_{p_k} x \}_{k < K}$.
For a total of $n_C$ classes, the union of all these feature defines 
a dictionary of size $M = K\, n_C$. They are linear combinations of the original
log scattering coefficients $\{ \phi_p x\}_p$. 
This dimension reduction 
can thus be interpreted as a last fully connected network layer, which outputs
a vector of size $M$. 
The parameter $M$ governs the bias versus variance trade-off. 
It can be adjusted from the decay of
the regression error of each $f_C$ or fixed a priori.
In classification experiments, $M$ is about $30$ times smaller than
the size of the original scattering dictionary.

The selected features are then provided to a Gaussian SVM classifier.
The variance of the Gaussian kernel
is set to the average norm of the scattering vectors, calculated from the 
training set. This large variance performs 
a relatively small localization in the feature space, but it
reduces classification errors as shown in Table 1. 

\section{Image Classification Results}\label{numeraics}

We compare the performance of a scattering network with 
state-of-the-art algorithms on CIFAR and Caltech datasets, which include
complex object classes, at different or fixed resolutions. 

Images of each databases are rescaled to become square images
of $2^{2d}$ pixels.
The scattering transform depends upon few parameters which are fixed a 
priori. 
The maximum scale of the scattering transform is set to $2^J = 2^{d-2}$.
Scattering coefficients are thus averaged over spatial domains covering
$1/4$ of the image width, and coefficients sampled over a spatial
grid of $4 \times 4$ points, a final down-sampling being performed without degrading
classification accuracies. This preserves 
some coarse localization information.
Coefficients are computed with Morlet wavelets having $L = 8$ orientations. 
The wavelet transform along these 
$L = 8$ angles are computed at a maximum scale 
$2^K = L/2$, which corresponds to a maximum angular variation of 
$\pi/2$. Indeed these object recognition problems do not involve larger rotation variability.
The resulting scattering
representation is nearly complete as previously explained.
It is computed independently 
along the 3 color channels YUV.
We apply a logarithm to separate illumination components.
The classifier is implemented by first reducing the dimensionality,
to $M = 2000$ feature vectors on CIFAR-10 for instance, with an orthogonal least square regression, and applying a Gaussian SVM. 
We use the same architecture and same 
hyperparameters for each dataset, 
apart from 
the number $M$ of selected coefficients, which increases proportionally to the
size of the scattering representation, which depends upon the image size.

Caltech-101 and Caltech-256 are two color image databases, with respectively
101 and 256 classes. They have
30 images  per class for training and the rest is used for testing.
Caltech images are rescaled to square images of $P=2^{2d} = 256^2$ pixels. 
Average per class classification results are reported 
with an averaging over 5 random splits. We removed the clutter class both from 
our training and testing set.

CIFAR are more challenging color image databases due to its high class variabilities, with
$60 000$ tiny colors images of $P= 2^{2d} = 32^2$ pixels. 
CIFAR-10 has 10 classes with 5000 training images per 
class, whereas 
CIFAR-100 has 100 classes with 500 training images per class.

Table 1 gives the classification accuracy for 
different scattering configurations,
on the datasets CIFAR-10 and Caltech-101. 
First order scattering coefficients are comparable to 
SIFT \cite{journals/pami/BrunaM13}, but are calculated over larger
neighborhoods. Second order scattering coefficients computed
with translated wavelets (no filtering along rotations) reduces
the error by $10\%$, which shows the importance of this complementary
information. Incorporating a wavelet filtering along rotations
as in Section \ref{scatnet},
leads to a further improvement of $4.5\%$ on Caltech-101 
and $1.2\%$ on CIFAR-10. Rotations produce larger pixel displacements on
higher resolution images. It may explain why improving sensitivity to
rotations plays a more important role on Caltech images, which are larger. 
Adding a feature reduction by orthogonal least square reduces the error by
$5.4\%$ on Caltech-101 and $0.7\%$ on CIFAR-10. The orthogonal least square
has a bigger impact on Caltech-101 because there are less 
training examples per class, so reducing the variance of the estimation
has a bigger effect.

\begin{table}
\begin{center}
\begin{tabular}{|l|c|l|}
\hline
Scattering & Caltech-101  & CIFAR-10 \\
\hline\hline

Trans., order 1 &59.8  &72.6\\
\hline
Trans., order 2&70.0& 80.3\\
Trans., order 2 + OLS&75.4 & 81.6\\
\hline
Roto-Trans., order 2 &  74.5& 81.5\\
Roto-Trans, order 2 + OLS & 79.9 & 82.3\\
\hline
\end{tabular}
\end{center}
\caption{Classification accuracy with 5 scattering configurations.
First a translation scattering up to order $1$, then up to order $2$, 
then withan Orthogonal Least
Square (OLS) feature reduction. Then a roto-translation scattering up to
order $2$, then with an OLS feature reduction.}
\end{table}

Tables 2,3,4,5 report the classification accuracy 
of a second order roto-translation scattering algorithm with an orthogonal
least square feature selection, for CIFAR and Caltech databases. It is compared
to state of the art algorithms, divided in four categories. 
``Prior'' feature algorithms apply a linear or an RBF type
classifier to a predefined set of features, 
which are not computed from training data. Scattering, 
SIFT and HOG vectors, or deep networks with random 
weights belong to this Prior class. 
``Unsup. Deep'' algorithms correspond to
unsupervised convolutional deep learning algorithms, whose filters are
optimized with non-labeled training data, before applying a linear classifier
or a Gaussian kernel SVM.
 ``Unsup. Dict.'' algorithms 
transform SIFT type feature vectors or normalized pixel patches,
with one, two or three successive sparse dictionaries computed by
unsupervised learning.
It may then be followed by a max-pooling operator
over a pyramid structure \cite{conf/cvpr/LazebnikSP06}.
To normalize unsupervised learning
experiments, we only consider results obtained without data augmentation. 
``Supervised'' algorithms compute feature or kernel representations,
which are optimized with supervised learning over labeled training data.
In this case, the training may be performed on a different databasis 
such as ImageNet, or may include a data augmentation by increasing 
the dataset with affine transformations and deformations.
Supervised deep convolution networks or supervised
kernel learning are examples of such algorithms.

Scattering gives better classification results than all 
prior feature classification on Caltech-101, as shown by Table 2.
Convolutional network with random filters on mono-CIFAR-10 (gray level CIFAR-10) have an accuracy of $53.2\%$ in \cite{conf/icml/SaxeKCBSN11}. Color
information improves classification results by at most 
$10\%$ on all algorithms, so it remains well below scattering accuracy.
No result is reported on CIFAR-100 using 
predefined ``prior'' feature classifiers.
Tables 2 and 4 shows that
scattering networks performs at least
as well as unsupervised deep convolutional architectures without
data augmentation on Caltech-101 and CIFAR-10. 
To our knowledge, no result with unsupervised deep convolutional network 
learning have been reported on Caltech-256.

State-of-the-art unsupervised classification results for Caltech, without data augmentation, are obtained with a Multipath-SC algorithm \cite{conf/cvpr/BoRF13}, 
which has 3 unsupervised encoding layers. Similar results are obtained with
Spatial Local Coding descriptors \cite{conf/accv/McCannL12} with a first layer of nearly SIFT descriptors followed by an unsupervised coding and multi-scale pyramidal pooling. 
Caltech-101 is an easier data basis because
it has a bias across classes, which is typically 
used by classifiers. This bias is removed from Caltech 256 which explains
why classifiers have a lower accuracy. The unsupervised classification algorithm reporting state of the art
results on CIFAR are different from the one on Caltech, which shows
that these figures must be analyzed with precaution. 
The scattering classifier gives comparable with all unsupervised
algorithms on CIFAR-10 and CIFAR-100.

Let us emphasize that we are using the same scattering representation, besides
image size adaptation, for Caltech and CIFAR databases. 
RFL \cite{conf/cvpr/JiaHD12} is the only
unsupervised learning algorithm which reports close to state of the
art results, both on Caltech and CIFAR data bases. 
RFL does not perform as well as a scattering on Caltech and CIFAR-100,
and slightly better on CIFAR-10. This illustrates the difficulty to have a single
algorithm which works efficiently on very different databases. 
We reported the result on CIFAR-100 from \cite{conf/cvpr/JiaHD12} via \cite{malinowski2013learning}. 

\begin{table}
\begin{center}
\begin{tabular}{|l|l|l|}
\hline
Method & Type  & Accuracy \\
\hline\hline
RotoTrans. Scat. & Prior & 79.9\\
Random Filters \cite{conf/iscas/LeCunKF10} & Prior &62.9\\
\hline

CDBN \cite{conf/icml/LeeGRN09}& Unsup. Deep & 65.4\\

\hline
M-HMP\cite{conf/cvpr/BoRF13}&Unsup. Dict& 82.5\\
SLC \cite{conf/accv/McCannL12}& Unsup. Dict & 81.0\\
Ask the locals \cite{conf/iccv/BoureauRBPL11}& Unsup. Dict  & 77.3\\
RFL \cite{conf/cvpr/JiaHD12}& Unsup. Dict & 75.3\\
\hline
CNN \cite{he2014spatial}& Supervised & 91.4\\
\hline
\end{tabular}
\end{center}
\caption{Results for different types of representations on Caltech-101.}
\end{table}

\begin{table}
\begin{center}
\begin{tabular}{|l|c|l|}
\hline
Method & Type  & Accuracy \\
\hline\hline

RotoTrans. Scat. & Prior & 43.6\\
\hline
M-HMP \cite{conf/cvpr/BoRF13}&Unsup. Dict & 50.7\\
SLC  \cite{conf/accv/McCannL12}& Unsup. Dict & 46.6\\
Ask the locals \cite{conf/iccv/BoureauRBPL11} & Unsup. Dict & 41.7\\
\hline
CNN \cite{journals/corr/ZeilerF13} & Supervised & 70.6\\
\hline
\end{tabular}
\end{center}
\caption{Results for different types of representations on Caltech-256.}
\end{table}

\begin{table}
\begin{center}
\begin{tabular}{|l|c|l|}
\hline
Method & Type  & Accuracy \\
\hline\hline
RotoTrans. Scat. & Prior & 82.3\\
\hline
RFL \cite{conf/cvpr/JiaHD12} & Unsup. Dict & 83.1\\
NOMP \cite{linstable} & Unsup. Dict & 82.9\\
\hline
LIFT \cite{conf/icml/SohnL12}& Unsup. Deep & 82.2\\
\hline
CNN \cite{lee2014deeply} & Supervised & 91.8\\
\hline
\end{tabular}
\end{center}
\caption{Results for different types of representations on CIFAR-10.}
\end{table}

\begin{table}
\begin{center}
\begin{tabular}{|l|c|l|}
\hline
Method & Type  & Accuracy \\
\hline\hline
RotoTrans. Scat. & Prior & 56.8\\

\hline
RFL \cite{conf/cvpr/JiaHD12} & Unsup. Dict & 54.2\\
NOMP \cite{linstable}& Unsup. Dict & 60.8\\
\hline
CNN \cite{lee2014deeply}& Supervised & 65.4\\
\hline
\end{tabular}
\end{center}
\caption{Results for different types of representations on CIFAR-100.}
\end{table}

The best classification results are obtained by supervised deep convolutional networks \cite{lee2014deeply,he2014spatial,journals/corr/ZeilerF13}. 
They improve non-supervised 
accuracy by about $10\%$  on CIFAR-10 or Caltech-101, $20\%$ on Caltech-256, but $5\%$ on CIFAR-100. The improvement on 
CIFAR-100 is smaller than on CIFAR-10 because there is only
500 samples per classes for supervised training, as opposed to 5000.
The Caltech data bases does not have enough training sample to train a
supervised deep network. We thus report classification results
obtained by the supervised Alex-network trained on ImageNet, 
to which is applied a linear SVM classifier which is trained on Caltech
\cite{journals/corr/GirshickDDM13}. 
Although this deep network was not trained on Caltech, it still
achieves the state of the art on this databases.  Experiments show that
if the training and testing image datasets are different, a supervised
deep network provides a feature vector having
a lower accuracy for classification, but this accuracy is not dramatically
reduced. It indicates that supervised deep classifiers are
learning generic image representations which are likely to capture more
complex geometric properties than unsupervised algorithms
or a roto-translation scattering transform.

A scattering transform is computed with convolutions along groups
of transformations which create important image variability. 
This paper concentrates on translations and rotations, but
it can be extended to any other group. Improving results
requires to consider other source of variabilities and invariants,
for example across color channels or across scales, which
are not recombined in this architecture.
Supervised deep neural networks do
apply non-linear transformations across color bands and scales.  
Computing wavelet and scattering 
transforms on arbitrary Lie groups or finite groups is not difficult
\cite{journals/corr/abs-1101-2286}. What is harder is to identify
the important group of variability for 
improving classification. It seems that supervised deep network classifiers
are able to identify them.

\section{Conclusion}

This work shows that feature vectors for image 
classification can be constructed from geometric image properties as opposed
to learning. A roto-translation scattering transform constructs a feature
vector providing joint information along multiple scales and multiple angles.
For complex object classification problems as in Caltech and CIFAR databases,
it considerably improves the performance of all existing
prior image descriptors, and
it yields comparable results with state of the art unsupervised deep learning,
and dictionary learning algorithms. 

Scattering networks do not
have instability properties as the ones 
observed for Alex-net \cite{journals/corr/SzegedyZSBEGF13},
because it applies contractive wavelet operators which are stable to deformations. 
However, deep neural networks with supervised training provide a clear improvement of average classification accuracy, compared to unsupervised learning and 
to this roto-translation scattering transform. 
This may indicate that they capture refined but important
geometric image properties. Understanding the nature of these properties is an
open challenge to further improve the performances of scattering representations.

\section*{Acknowledgments}
This work is funded by the ERC grant InvariantClass 320959 and via a grant for Phd Students of the Conseil r\'egional d'\^Ile-de-France (RDM-IdF).
{\small
\bibliographystyle{ieee}
\bibliography{my_bib}
}

\end{document}